# Feature Enhancement in Visually Impaired Images

**Madhuri Suthar[1*], Mohammad Asghari[1] and Bahram Jalali[1,2,3,4]**

[1]*Department of Electrical Engineering, University of California Los Angeles, Los Angeles, California, USA*

[2]*California NanoSystems Institute, Los Angeles, California, USA*

[3]*Department of Bioengineering, University of California Los Angeles, Los Angeles, California, USA*

[4]*Department of Surgery, David Geffen School of Medicine, University of California Los Angeles, Los Angeles, California, USA*

[*]*Corresponding author email: madhurisuthar@ucla.edu*

## Abstract

One of the major open problems in computer vision is detection of features in visually impaired images. In this paper, we describe a potential solution using Phase Stretch Transform, a new computational approach for image analysis, edge detection and resolution enhancement that is inspired by the physics of the photonic time stretch technique. We mathematically derive the intrinsic nonlinear transfer function and demonstrate how it leads to (1) superior performance at low contrast levels and (2) a reconfigurable operator for hyper-dimensional classification. We prove that the Phase Stretch Transform equalizes the input image brightness across the range of intensities resulting in a high dynamic range in visually impaired images. We also show further improvement in the dynamic range by combining our method with the conventional techniques. Finally, our results show a method for computation of mathematical derivatives via group delay dispersion operations.

## Introduction

Feature detection in images plays a critical role in the field of computer vision for solving problems associated with object recognition, image registration, content-based image retrieval and deep learning [1-3]. Prior works for improving the feature detection in images have focused on use of grey level statistics of the image [1] and on application of edge detection methods [4]. Color distinctiveness and color models [5, 6] and scale selections [7] in images have also been exploited for enhancing the feature detection. The main goal of feature detection is to classify objects more accurately and at the same time be robust to varying viewing conditions that include changes in illumination or environment conditions, object orientation as well as zoom factor of the camera. Environmental conditions can severely impair detection and localization of objects in images. For instance, under foggy condition, acquired images suffer from visual impairments that reduce contrast, cause blur and noise leading to lower resolution and lower contrast [8, 9]. This constitutes a major bottleneck for many computer vision applications, in particular, for autonomous vehicles. While the emerging imaging technologies such as High Dynamic Range (HDR) hold promise to solve feature detection problems in the computer vision but their slow frame rate pose a challenge in self-driven cars, autonomous robotics and other real-time applications.

The Phase Stretch Transform (PST) was recently introduced as a computational approach to signal and image processing [10,11]. PST is a physics-based algorithm that has its roots in photonic time stretch technique [12-15], a method for real-time measurements of ultra-fast events and one that has enabled the discovery of optical rogue waves [16], observation of relativistic electron structure [17], label-free cancer cell detection with record accuracy [18] and optical data compression [19]. The algorithm mimics the propagation of electromagnetic waves through a diffractive medium with engineered 3D dispersive property (refractive index) [10,11]. PST can be applied to both digital images as well as time series data [20] and has been used for edge detection in biomedical images [11, 21-22] and Synthetic Aperture Radar (SAR) images [23]. PST has also been applied for resolution enhancement in super-resolution localization microscopy for imaging of single molecule [24]. The algorithm has been open sourced on GitHub and Matlab Central File Exchange [25].

We first show that PST has an inherent equalization ability that gives a response ideal for feature detection in low contrast regimes of visually impaired images caused by the fog. To do this, we apply the algorithm on two road traffic images under fog as shown in Fig 1. It shows how the algorithm could significantly improve the feature detection by outperforming the conventional edge detection methods based on derivative of the image. The method based on derivative is unable to capture details with small contrast variations in the brighter low resolution areas where as our technique clearly shows the contrast changes in these visually impaired regions. The warp and strength parameters of PST kernel as described in [10, 11] for the feature detection in these images are 22 and 500, respectively. As we will show in our mathematical formulations, this property emerges because PST's transfer function is a reconfigurable operator. Later, we examine the superior performance of PST at low light levels and its application to HDR images.

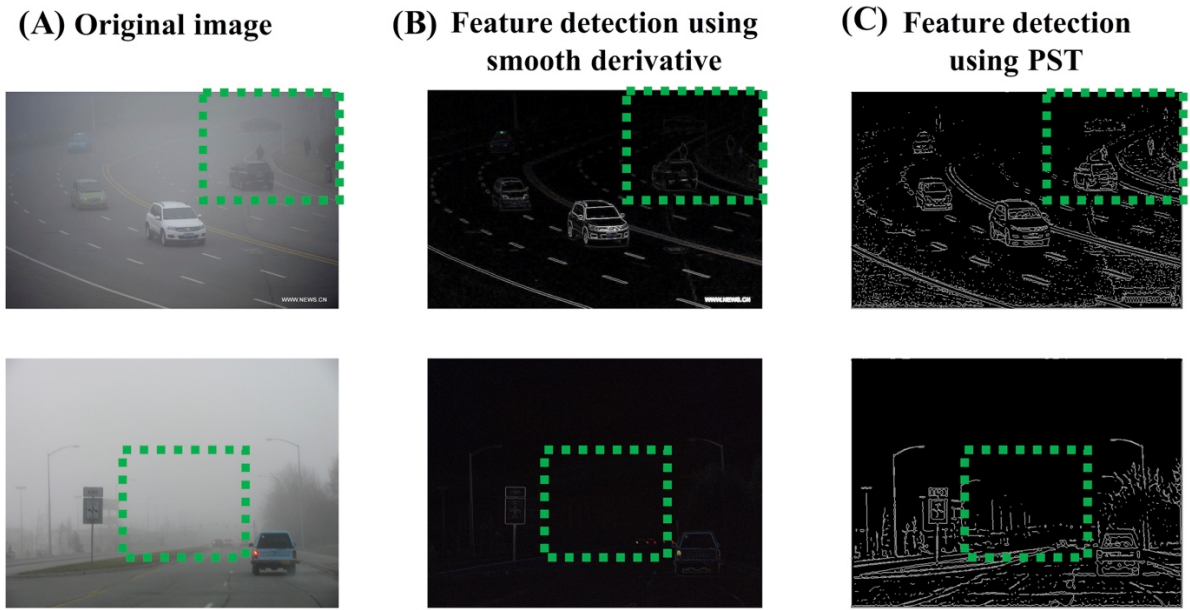

**(A) Original image**   **(B) Feature detection using smooth derivative**   **(C) Feature detection using PST**

**Fig 1. Comparison of feature detection using conventional derivative operator to the case of feature detection using PST in case of visually impaired images under fog conditions.**
Original images are shown in (A). Results of feature detection using conventional derivative operator and PST operator are shown in (B) and (C), respectively. Note that the conventional derivative operator misses to efficiently visualize the small contrast changes in visually impaired regions of the images (shown in green dashed boxes). However, PST can clearly show the contrast changes in the low resolution regions (shown in green dashed boxes) due to its reconfigurable mechanism that enables highlighting features over a wide dynamic range. The strength of features detected using PST over both low and high resolution regions of the images is consistent unlike derivative operator.

# Mathematical foundations of Phase Stretch Transform

We prove the superior performance of Phase Stretch Transform in the low contrast regime by deriving closed-form analytical expressions for its transfer function. Mathematical results reveal that the transform has an intrinsic intensity equalization property leading to high dynamic range performance. Analytical results are supported by numerical simulation confirming the dynamic range enhancement. We define the stretch operator $\mathbb{S}\{\ \}$ as follows

$$E_o[x,y] = \mathbb{S}\{E_i[x,y]\} \triangleq \text{IFFT2}\left\{\widetilde{K}[u,v] \cdot \widetilde{L}[u,v] \cdot \text{FFT2}\{E_i[x,y]\}\right\} \qquad (1)$$

and $E_o[x,y]$ is a complex quantity defined as,

$$E_o[x,y] = |E_o[x,y]|\, e^{j\theta[x,y]} \qquad (2)$$

In the above equations, $E_i[x,y]$ is the input image, x and y are the spatial variables, FFT2 is the two dimensional Fast Fourier Transform, IFFT2 is the 2-dimensional Inverse Fast Fourier Transform and u and v are spatial frequency variables. The function $\widetilde{K}[u,v]$ is called the warped phase kernel and the function $\widetilde{L}[u,v]$ is a localization kernel implemented in frequency domain. Here for simplicity, we assume $\widetilde{L}[u,v] = 1$. PST operator is defined as the phase of the Warped Stretch Transform output as follows [10],

$$\text{PST}\{E_i[x,y]\} \triangleq \sphericalangle\left\{\mathbb{S}\{E_i[x,y]\}\right\} \qquad (3)$$

where $\sphericalangle\langle\cdot\rangle$ is the angle operator.

Without the loss of generality and in order to keep the notations manageable in what follows, we consider operation of PST on 1D data, i.e.,

$$\text{PST}\{E_i[x]\} = \sphericalangle\{E_o[x]\} = \sphericalangle\left\langle \text{IFFT}\left\{\widetilde{K}[u] \cdot \text{FFT}\{E_i[x]\}\right\}\right\rangle \qquad (4)$$

The warped phase kernel $\widetilde{K}[u]$ is described by a phase function with nonlinear dependence on frequency, u,

$$\widetilde{K}[u] = e^{j\cdot\varphi[u]} \qquad (5)$$

By using the Taylor expansion for the phase term in the kernel $\widetilde{K}[u]$ we have,

$$\widetilde{K}[u] = e^{\left(j\sum_{m=2}^{M}\frac{\varphi^{(m)}}{m!}u^m\right)} \qquad (6)$$

where $\varphi^{(m)}$ is the $m^{th}$-order discrete derivative of the phase $\varphi[u]$ evaluated for u=0 and values of m are even numbers. PST phase term $\varphi[u]$ only contains even-order terms in its Taylor expansion due to even symmetry requirement for the phase term $\varphi[u]$ for proper operation of PST [20]. Using the warped phase kernel described in Eq. (6), output complex-field, $E_o[x]$, can be calculated as follows,

$$E_o[x] = IFFT\{\widetilde{E}_i[u] \times \widetilde{K}[u]\}$$

$$= IFFT\left\{\widetilde{E}_i[u] \times e^{\left(j\sum_{m=2}^{M}\frac{\phi^{(m)}}{m!}u^m\right)}\right\} \qquad (6)$$

where $\widetilde{E}_i[u]$ is the discrete Fourier transform of the input. Simulation show that PST works best when the applied phase is small. Under these conditions, using small value approximation, the exponential term in Eq. (7) can be simplified to,

$$E_o[x] = IFFT\left\{\widetilde{E}_i[u] \times \left[1 + j\left(\sum_{m=2}^{M}\frac{\phi^{(m)}}{m!}u^m\right)\right]\right\} \qquad (8)$$

$$\rightarrow \quad E_o[x] \approx \left[1 \times E_i[x] + j\sum_{m=2}^{M}\frac{(-1)^{m/2}\phi^{(m)}}{m!\ (2\pi)^m}E_i[x]^{(m)}\right] \qquad (9)$$

where $E_i[x]^{(m)}$ is the $m^{th}$-order discrete derivative of the input $E_i[x]$. Since the input is a real quantity, the output phase can be calculated as,

$$PST\{E_i[x]\} = \angle\{E_o[x]\} \approx tan^{-1}\left\{\frac{\sum_{m=2}^{M}\frac{(-1)^{m/2}\phi^{(m)}}{m!\ (2\pi)^m}E_i[x]^{(m)}}{E_i[x]}\right\} \qquad (10)$$

Finally, since the phase is restricted to small values, Eq. (10) can be simplified to,

$$PST\{E_i[x]\} \approx \frac{\sum_{m=2}^{M}\frac{(-1)^{m/2}\phi^{(m)}}{m!\ (2\pi)^m}E_i[x]^{(m)}}{E_i[x]} \qquad (11)$$

We see that the transfer function consists of summation of even order derivatives in the numerator divided by the amplitude (brightness) in the denominator. The numerator extracts a hyper dimensional set of features that corresponding to different measures of the curvature of the edge. The denominator renders the response nonlinear in such a way that low light levels are enhanced. These results were obtained for a general phase kernel and for small values of phase in the PST kernel. We now consider two additional scenarios revealing further insight into the unique properties of our transform.

**Case 1:** We consider the Phase Kernel $\widetilde{K}[u] = u^2$ as a quadratic function of frequency variable u. Under this condition, using small phase approximation as used before, the exponential term in Eq. (8) can be simplified to,

$$E_o[x] = \text{IFFT}\{\ \widetilde{E}_i[u] \times [1 + j\,(u^2)]\ \} \qquad (12)$$

$$\rightarrow \qquad E_o[x] \approx \left[1 \times E_i[x] - j\,\frac{1}{(2\pi)^2} * \frac{d^2 E_i[x]}{dx^2}\right] \qquad (13)$$

Finally, for calculating the phase of the complex output we assume it to be restricted to small values. Therefore, phase of the output in Eq. (13) can be simplified to,

$$\text{PST}\{\ E_i[x]\} = \measuredangle\ E_o[x] \approx \frac{\frac{-1}{(2\pi)^2} * \frac{d^2 E_i[x]}{dx^2}}{E_i[x]} \qquad (14)$$

**Case 2:** We consider here, the same Phase Kernel $\widetilde{K}[u] = u^2$ as a quadratic function of frequency variable u as discussed in case 1. However, we remove the use of small phase approximation. The exponential term in Eq. (8) now leads to,

$$E_o[x] = \text{IFFT}\{\ \widetilde{E}_i[u] \times [\cos(u^2) + j\sin(u^2)]\ \} \qquad (15)$$

Expanding the sine and cosine terms using Euler expansion up to third order and then applying small value approximation to the complex output of Eq. (12) will result in the complex PST as shown below

$$\text{PST}\{\ E_i[x]\} = \measuredangle\ E_o[x] \approx \frac{\frac{-1}{(2\pi)^2} * \frac{d^2 E_i[x]}{dx^2} + \frac{1}{3!(2\pi)^6} * \frac{d^6 E_i[x]}{dx^6} - \frac{1}{5!(2\pi)^{10}} * \frac{d^{10} E_i[x]}{dx^{10}}}{E_i[x] - \frac{1}{2!(2\pi)^4} * \frac{d^4 E_i[x]}{dx^4} + \frac{1}{4!(2\pi)^8} * \frac{d^8 E_i[x]}{dx^8}} \qquad (16)$$

The closed-form expression presented in Eq. (11) relates the output to the input in the case of arbitrary phase kernel with small phase approximation. To give an example, the core functionality of the PST as a feature detector can be understood by closed-form expression shown in Eq. (11). The output of the PST operator is related directly to the even-order derivatives of the input with weighting factors of $\frac{(-1)^{m/2}\,\phi^{(m)}}{m!\,(2\pi)^m}$. Each derivatives detects a different feature in the input. Thus, the weighting factors can be designed to select features of interest. In other words, our transform is a reconfigurable operator that can be tuned to emphasize different features in an input image.

The important observation from Eq. (11) is that, the output is inversely related to the input brightness level and this is valid for small phase approximation. Therefore, for the same contrast level, the output is larger in darker low light level parts of the image. This crucial property inherent in PST equalizes the input brightness and allows for more sensitive feature detection and enhancement. Brightness level equalization is a well-studied method to improve feature detection algorithms in High Dynamic Range (HDR) images (see [26] for an example). One technique for brightness level equalization in images is a log function applied to the input before feature detection. The log function has a higher gain for lower brightness input which equalizes the brightness and results in more efficient feature detection. Fortunately, PST operator has a built-in logarithmic behavior which gives it excellent dynamic range. However, this does not completely describe the transform. As observed in Eq. (16), our transform outputs a hyper-dimensional feature set for classification. These results also show a method for computation of mathematical derivatives via group delay dispersion operations.

## Simulation results

In this section we present some simulation results that confirm the closed-form expression derived in previous section. We also show examples of operation of PST on HDR images, supporting the new theory developed above. In the first example, we simulate the output for a given 1D data and compare it to the output estimated by the mathematical expression derived in Eq. (11). The phase kernel $\varphi[f_x]$ designed for the PST operator is shown in Fig 2A. The warp and strength parameters of the kernel as described in [10,11] are 12.5 and 4000, respectively. The phase and its derivative are shown in Fig 2A and Fig 2B respectively. The input is shown in Fig 2C. Numerically simulated output is compared to the output estimated by Eq. (11) in Fig 2D using red-solid and blue-dotted lines, respectively. It is evident that the simulations confirm the accuracy of the closed-form analytical model of our algorithm.

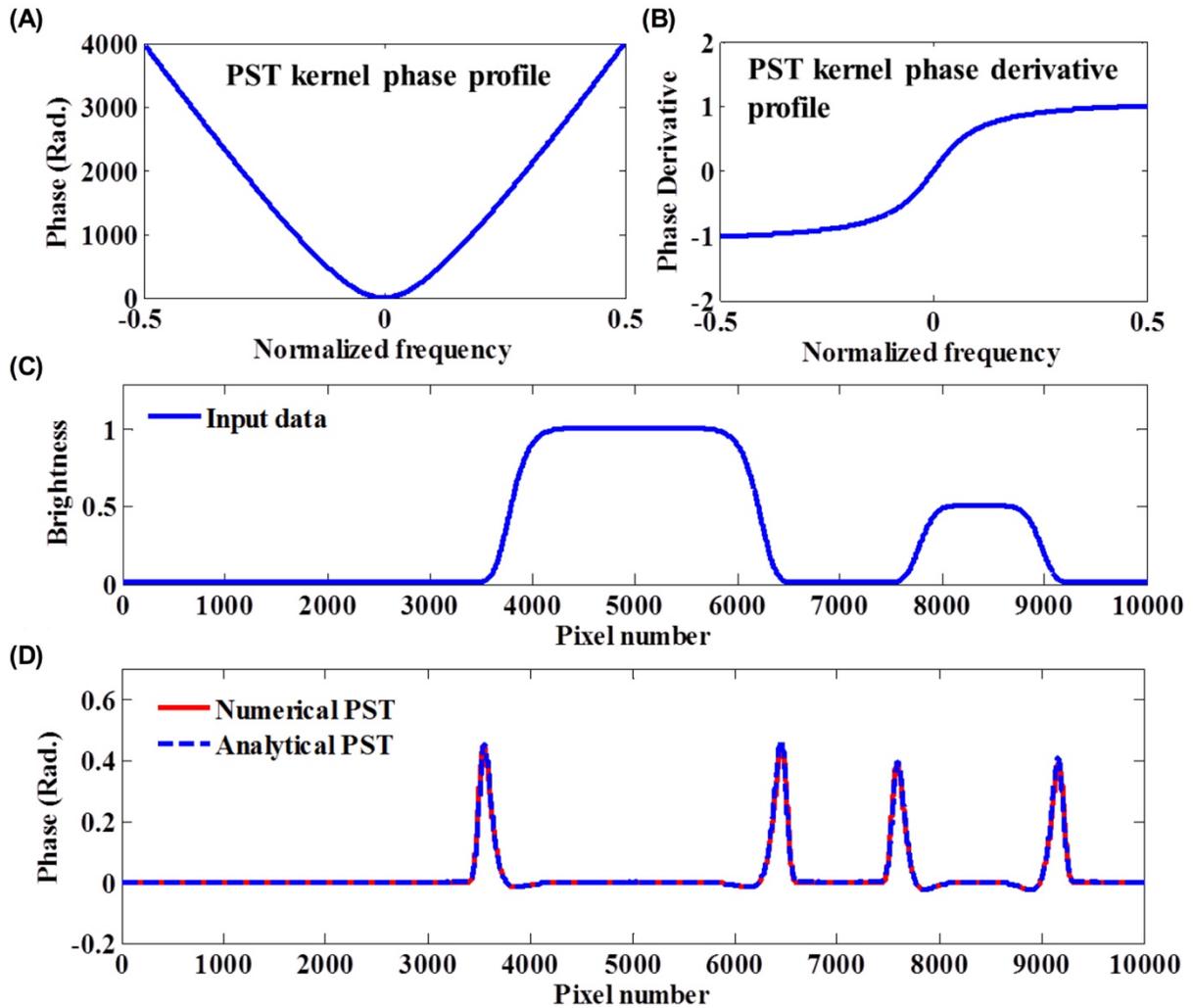

**Fig 2. Comparison of numerically simulated output of PST algorithm with the output given by the closed-form analytical expression derived in Eq. (11).**
The phase kernel and the corresponding phase derivative profile are shown in (A) and (B), respectively. The input 1D brightness data is shown in (C). Numerically calculated output data is compared to the output data estimated by Eq. (11) in (D) using red-solid and blue-dotted lines, respectively. Simulation results confirm the accuracy of the closed-form analytical model in Eq. (11).

In the next example, we evaluate the effect of PST on features with different contrast levels at a fixed brightness level and compare it to the case of using the conventional technique of differentiation to detect features in the same input. The warp and strength parameters used for the PST operator are 12.15 and 0.48, respectively. The input was designed to have different contrast level change at fixed brightness level, see Fig 3A. Numerically simulated output a using PST is compared to the output using differentiation in Fig 3B. As expected, the output of the differentiator is linearly related to the contrast level and is insensitive to the brightness level. However, the relation of the PST output to contrast level at fixed brightness level is nonlinear. This effect is due to the brightness level equalization feature of PST described in Eq. (11).

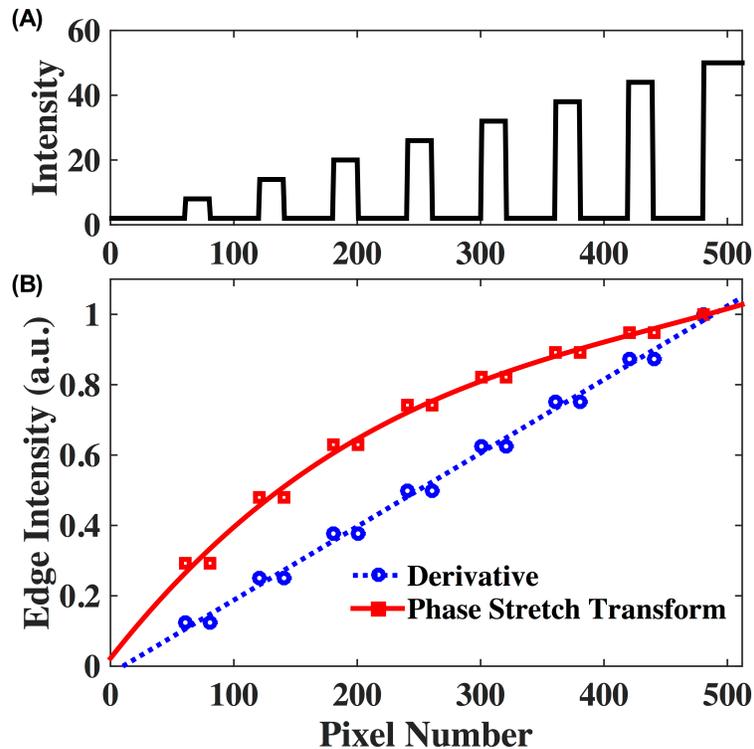

**Fig 3. Effect of PST on features with different contrast levels at fixed brightness level compared to the output by differentiation.**

The input was designed to have different contrast level change at fixed brightness level, shown in (A). Numerically simulated output using PST is compared to the output using differentiation in (B) using red-solid and blue-dotted lines, respectively. The output of the differentiator is linearly related to the contrast level and is insensitive to the brightness. However, the dependence of the PST output to contrast at fixed brightness is nonlinear. This effect is due to the equalization mechanism of PST described in Eq. (11).

Fig 4 shows an example of using PST for feature enhancement in 14 bits HDR image. The image has sharp features in the dark regions, as shown in the dashed box. The image also has features in a region that the brightness slowly decreases, as shown in the sloid box. Here we compare the feature detection using the derivative operator with that using PST. The derivative operator was implemented from native smooth derivative function. For fair comparison, both methods use the same localization kernel with sigma factor of 2. The warp and strength parameters used for the PST operator are 12.15 and 0.48, respectively. Results of feature detection using smooth derivative operator and PST operator are shown in *Fig 4*B  and *Fig 4*C, respectively. The derivative operator is unable to unveil the small contrast changes in dark areas of the image, see dashed box in Fig 4B . However, PST extracts more information on the contrast changes in dark areas due to its natural equalization mechanism, see dashed box in Fig 4C. It also can be observed that the intensity of detected edges in the case of smooth derivative is related linearly to the brightness level of the original image, compare solid box areas in Fig 4A and Fig 4B. In contrast, PST has automatically equalized the brightness level in the solid box region in the image and outputs relatively fixed features intensity for that region, see solid box in Fig 4C. We note that PST has missed to visualize features in very bright areas in the image. This is because of the inverse dependence on brightness as shown in Eq. (11). This issue can be mitigated by setting a maximum threshold for detected features or by equalizing the image brightness before passing through PST operator.

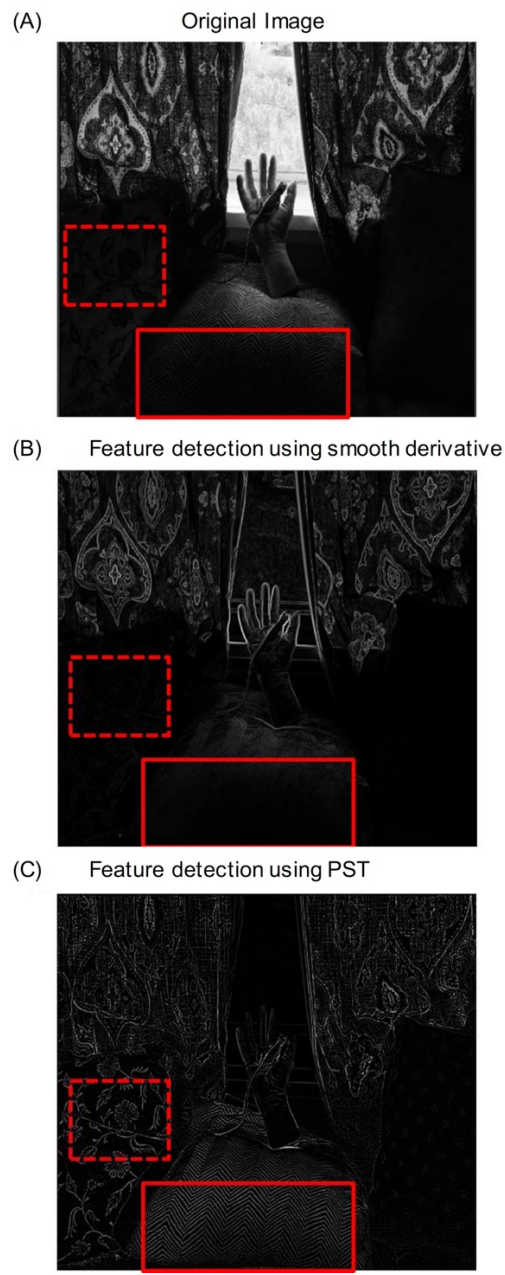

(A) Original Image

(B) Feature detection using smooth derivative

(C) Feature detection using PST

**Fig 4. Comparison of feature detection using smooth derivative operator to the case of feature detection using PST.**

Original image is shown in (A). Smooth derivative operator is unable to efficiently visualize the small contrast changes in dark areas of the image. However, PST can clearly show the contrast changes in dark areas due to its natural equalization property.

To better explain, the role of PST for feature detection in low-light-level and high-light-level regions, we use a single line scan of the HDR image shown previously in Fig 4. The blue box in the Fig 5 demonstrates the response of PST to low-light-level regions where it outperforms conventional derivative operator. Similarly, for high-light-level regions of the image (shown in green and purple box in the Fig 5), PST outperforms when the contrast is low (shown in green box in the Fig 5). In contrast, the conventional derivative operator response is dominating only under high contrast (shown in purple box in the Fig 5).

**(A)   Input Line Scan**

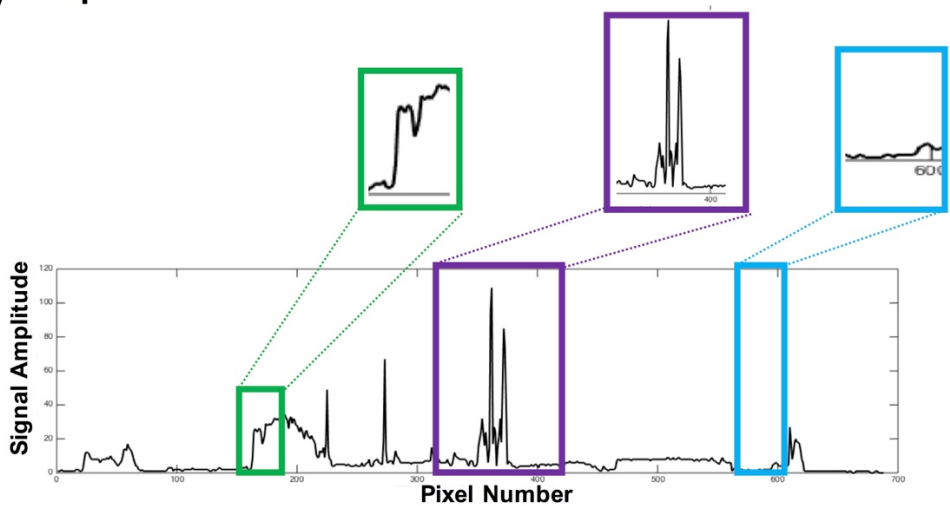

**(B)   Feature Detection using PST and conventional derivative operator**

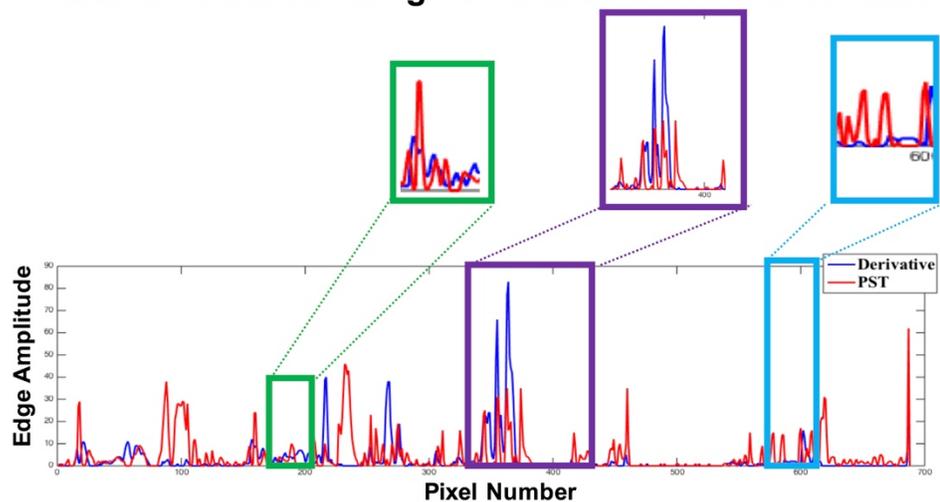

**Fig 5. A line scan corresponding to Row 524 from the image previously shown in Fig 4 comparing feature detection using conventional derivative operator and PST operator under low-light-level and high-light-level conditions.**

Original input line scan is shown in (A). Feature detection of this line scan using derivative and PST operators are shown in (B). The blue box highlights that the feature detection response of PST is much larger than the derivative operator under low-light-level conditions. The green and purple box demonstrates the response of PST and derivative operator for feature detection under high-light-level conditions. While PST enhances features under low contrast also under high-light-levels (see green box) unlike derivative operator which identifies features only under high contrast (see blue box).

To further enhance the dynamic range, we introduce a hybrid system that selects the best edge response from both the PST and the conventional edge detection filter. As shown in Fig , the hybrid system combines the edge response from PST and conventional derivative operator, providing edge detection in both low-light and high-light levels under low and high contrast variations.

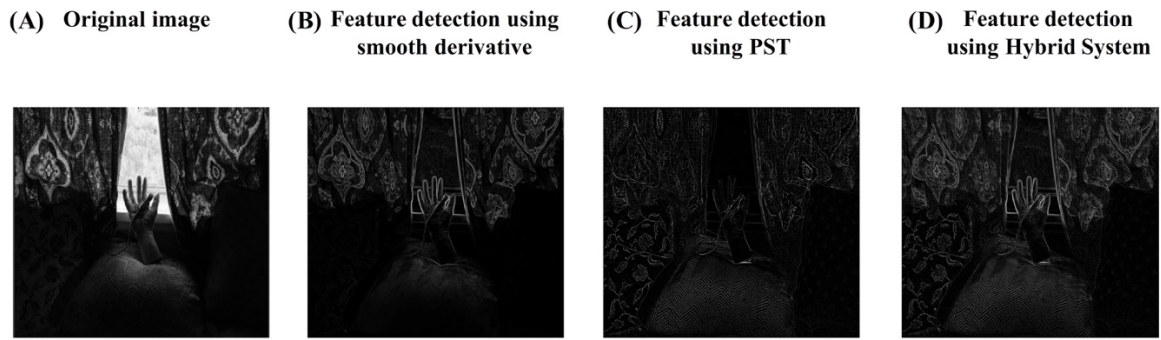

**Fig 6. Hybrid system for feature detection using derivative operator and PST.**
The Original image is shown in (A). Results of feature detection using smooth derivative operator and PST operator are shown in (B) and (C), respectively. And the output of hybrid system is shown in (D). Note that in (D), the strength of the detected features in both the high-light-level and low-light-level regions is same. The hybrid system selects the detected features in the darker regions using PST operator and in the brighter regions using smooth derivative operator providing a wider range of dynamic operation.

# Conclusions

In this paper, we presented a new method for edge detection in visually impaired images using a new mathematical transform inspired by the physics of photonic time stretch. We showed via analytical derivation and numerical simulations that the so called Phase Stretch Transform equalizes the input brightness resulting in a high dynamic range in feature detection and acts as a hyper-dimensional feature set classifier. Furthermore, our results show a method for computation of mathematical derivatives via group delay dispersion operations.